\documentclass[letterpaper, 10 pt, conference]{ieeeconf}  

\IEEEoverridecommandlockouts                              

\usepackage{graphics} 
\graphicspath{ {./} }

\usepackage{amsmath,amsthm}
\usepackage{amssymb}  
\usepackage{amsfonts} 
\usepackage{graphicx,float,wrapfig}
\usepackage{cite}
\usepackage{bm}
\usepackage{soul,color}

\makeatletter
\let\NAT@parse\undefined
\makeatother
\usepackage[numbers,sort&compress]{natbib}
\usepackage{balance}

\usepackage[linesnumbered,ruled]{algorithm2e}
\usepackage{url}
\usepackage{booktabs}
\usepackage[table,xcdraw,x11names]{xcolor}

\usepackage{geometry}
\usepackage{subcaption}
\usepackage[font=small]{caption}
\usepackage{multirow}

\usepackage{siunitx}
\usepackage{mathtools}
\usepackage{xspace}

\newcommand{\mPerSecN}[0]{\si[per-mode=symbol]{\si{\meter}\per\si{\second}}}

\newcommand{\mPerSecCuN}[0]{\si[per-mode=symbol]{\si{\meter}\per\si{\second}\cubed}}

\usepackage{hyperref}

\let\labelindent\relax
\usepackage{enumitem}

\newtheorem*{problem*}{Problem}

\title{\LARGE \bf
Driving in Dense Traffic with Model-Free Reinforcement Learning
}

\author{Dhruv Mauria Saxena$^{1}$, Sangjae Bae$^{2}$, Alireza Nakhaei$^{3}$, Kikuo Fujimura$^{3}$, Maxim Likhachev$^{1}$
\thanks{This work was funded by Honda Research Institute USA, Inc.}
\thanks{$^{1}$ Carnegie Mellon University, USA.
        {\tt\small \{dsaxena, mlikhach\}@andrew.cmu.edu}}%
\thanks{$^{2}$ University of California, Berkeley, USA.
        {\tt\small sangjae.bae@berkeley.edu}}%
\thanks{$^{3}$ Honda Research Institute USA, Inc.
        {\tt\small \{anakhaei,kfujimura\}@honda-ri.com}}%
}

\begin{document}
\maketitle
\thispagestyle{empty}
\pagestyle{empty}

\begin{abstract}
Traditional planning and control methods could fail to find a feasible trajectory for an autonomous vehicle to execute amongst dense traffic on roads. This is because the obstacle-free volume in spacetime is very small in these scenarios for the vehicle to drive through. However, that does not mean the task is infeasible since human drivers are known to be able to drive amongst dense traffic by leveraging the cooperativeness of other drivers to open a gap. The traditional methods fail to take into account the fact that the actions taken by an agent affect the behaviour of other vehicles on the road. In this work, we rely on the ability of deep reinforcement learning to implicitly model such interactions and learn a continuous control policy over the action space of an autonomous vehicle. The application we consider requires our agent to negotiate and open a gap in the road in order to successfully merge or change lanes. Our policy learns to repeatedly probe into the target road lane while trying to find a safe spot to move in to. We compare against two model-predictive control-based algorithms and show that our policy outperforms them in simulation.
\end{abstract}

\section{Introduction}\label{sec:intro}
\begin{figure}[t]
    \centering
    \includegraphics[width=0.8\columnwidth]{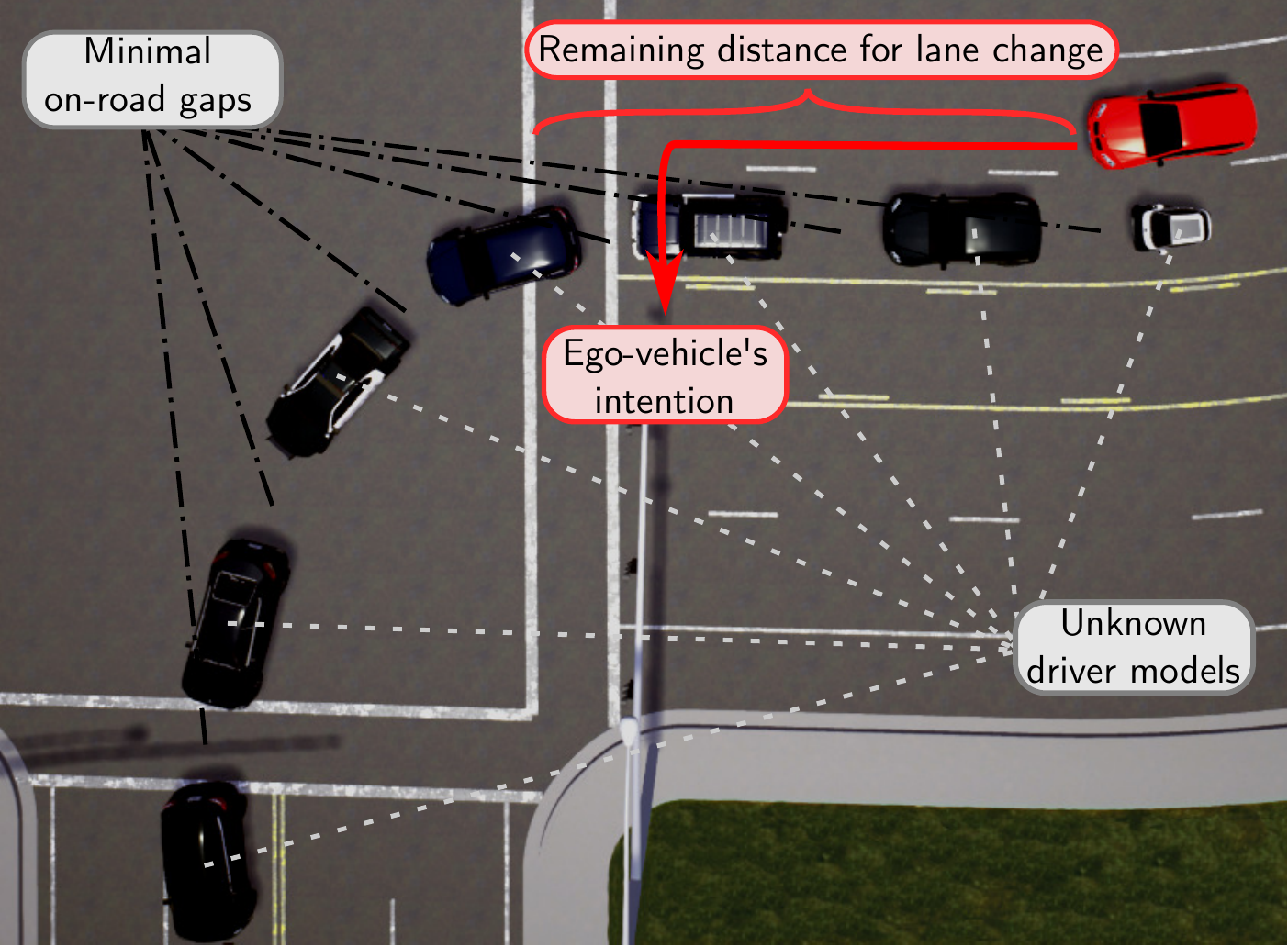}
    \caption{\textbf{Motivating Example:} The ego-vehicle (in \textcolor{Red1}{red}) wants to change lanes in order to make a legal left turn. Road rules restrict the distance available for this lane change. There is also dense traffic on the road. The driver models of \textit{all} other vehicles are unknown to the ego-vehicle (in terms of their cooperativeness at the very least).}
    \label{fig:example}
    \vspace{-10pt}
\end{figure}

Since the 2007 DARPA Urban Challenge~\cite{DARPAUrban}, autonomous vehicles have transitioned from being tested in well-structured environments under complete supervision and control, to being tested on actual roads in the real-world amongst human drivers. This progress comes together with several challenges that must be addressed if autonomous vehicles are to share the road with human driven vehicles one day. One of these is the fact that driving on roads is inherently an interactive exercise, i.e. actions taken by an autonomous vehicle affect other nearby vehicles on the road and vice versa~\cite{CoopPlanning}. This is especially apparent in dense traffic where any goal-directed behaviour must rely on some level of cooperation between various agents on the road in order to achieve the desired goal. For example, consider the motivating example of this work from Fig.~\ref{fig:example}. The goal for the \textcolor{Red1}{red} \textit{ego-vehicle}\footnote{We refer to the vehicle or agent of interest as the ego-vehicle. In the remainder of this paper, this will be the vehicle we control on the road.} is to change into the left lane before the intersection so that it can make a legal left turn. However, the dense traffic on the road makes it necessary for the ego-vehicle to convince a vehicle in that lane to give it room in order to successfully change lanes. In the remainder of this paper, we refer to the finite distance available to the ego-vehicle as the distance to a \textit{deadend}.

Traditional planning and control techniques might fail to find reasonable solutions in these scenarios due to the minimal volume of obstacle-free space~\cite{levinson2011towards,katrakazas2015real,PadenCYYF16}. As a result, the only feasible, collision-free solution for the autonomous vehicle might be to stay stationary which incurs other costs (such as increased time to achieve the goal and decreased average speed)~\cite{FRP}. It is possible to augment these methods to take predictions about the trajectories of other vehicles into account~\cite{SearchBasedAD,TrajOptAD}, but incorporating complex interactions between multiple vehicles as part of these techniques is still an open research question. In this work, we present a model-free reinforcement learning based solution that does not need any explicit model of inter-vehicle interactions, yet leverages these interactions to learn behaviours that successfully accomplish the lane change task from Fig.~\ref{fig:example}. We state our problem of interest succinctly below.
\begin{problem*}
    Successfully execute a safe and comfortable merge into dense traffic.
\end{problem*}
In order to avoid the ``freezing robot problem'' in dense traffic, even perfect predictions of future trajectories of other agents is not enough. We must rely on implicit or explicit models of \textit{joint} interactions between the agents in the scene~\cite{FRP}. For autonomous driving, since we cannot reliably know the behaviour of other drivers,
we choose to implicitly learn this joint interaction behaviour by utilising the ability of deep reinforcement learning to learn complex policies from data~\cite{franccois2018introduction}.

Our main contributions in this work are:
\begin{itemize}
    \item a challenging benchmark for driving in dense traffic (Section~\ref{sec:benchmark}).
    \item a continuous state and action space, policy gradient-based deep reinforcement learning solution for the benchmark (Section~\ref{sec:network}).
    \item a design specification for agent observations and the reward function for this task (Sections~\ref{sec:obs} and~\ref{sec:reward}).
\end{itemize}

\section{Related Work}\label{sec:litreview}
In recent years, there has been a considerable focus on using machine learning techniques for autonomous driving. These have been used for policies learned via behaviour cloning and reinforcement learning. A recent survey by Schwarting, Alonso-Mora, and Rus~\cite{survey} is a great resource for motion planning for autonomous driving in genral and covers literature on both learning-based methods and traditional methods (search-based~\cite{AjanovicLSSH18}, sampling-based~\cite{MaXKZMZ15}, or optimisation-based~\cite{XuWDZZ12} planning techniques). We focus on the former category below as it is most relevant to our work.


\subsection{Behaviour Cloning}\label{sec:bc}
There has been work on using neural networks to control steering commands of autonomous vehicles since the late 1980s~\cite{Pomerleau-1989-15721}. Recent developments in data processing and computational resources have caused a shift towards deep learning-based methods~\cite{BojarskiTDFFGJM16}. Both approaches rely on behaviour cloning to learn a mapping from raw images directly to steering angle commands using a dataset of human-driven trajectories. Behaviour cloning has been used to learn a mapping from input observations (raw sensor data or processed information) to control policies in several ways~\cite{ChenDeepDriving,KueflerMWK17,CodevillaMLKD18,AminiIROS,HenaffCL19}. These approaches differ in terms of the input observation, action space, policy parameterisation, or additional outputs (like uncertainty estimates for the chosen action), but they all still rely on labeled datasets. We also train a policy end-to-end in our work, but reinforcement learning has no supervisory label as in these approaches.

\subsection{Reinforcement Learning}\label{sec:rl}
Given its early success on other continuous control tasks~\cite{lillicrap2015continuous}, reinforcement learning has received considerable attention for autonomous driving~\cite{Shalev-ShwartzS16a,ChenDRL,HoelDRL,WolfHWBHDZ17}. The complexity of the task being solved varies greatly across the literature, along with the reinforcement learning algorithms used and parameterisation of the control policy. Q-Learning has been used to control the steering angle for a single vehicle on a road without traffic~\cite{WolfHWBHDZ17}. In~\cite{HoelDRL}, reinforcement learning is used to learn a value function which in turn is used to guide a Monte Carlo Tree Search for action selection. In contrast to these approaches, we use reinforcement learning to learn a low-level, continuous-control policy for driving amongst dense traffic. The continuous control aspect has been addressed previously~\cite{ChenDRL}, however the scenario they consider does not necessitate complex interactions with other vehicles in order to successfully complete the task. The \textit{double-merge} task from~\cite{Shalev-ShwartzS16a} does require vehicles to interact with others in order to achieve its goal, but they focus on learning higher-level tactical decisions as opposed to low-level controls.
More recently,~\cite{YepingHRI} used reinforcement learning to solve problems similar to the one we consider. However their action space was much simpler than ours (five discretised values of acceleration), there was very little traffic on the road, and the road geometry only accommodated a fixed merge \textit{point} as opposed to the more realistic case of a finite distance for the task (lane change or merge) as in our work.

Explicitly modeling human interactions could help elicit better behaviours from our learned policy in terms of safe driving on the road, passenger comfort and interpretability of agent behaviours. There is promising work in this direction~\cite{SadighSSD16,FisacBSSSD19}, however we choose to keep our work model-free since in our scenario, we would need to capture interactions with several vehicles at every timestep, which could be computationally intractable even if there was a reasonable model.


\section{Approach}\label{sec:approach}
Continuous control reinforcement learning algorithms
use policy gradient optimisation to directly learn a policy over the state space, which can be easier than first learning a value function and using it to derive a policy. Briefly, these algorithms maximise the following objective via gradient ascent,
\begin{align*}
    \nabla_\theta &J(\theta) = \\
    &\mathbb{E}_{\tau \sim \pi_\theta(\tau)}\left[ \left(\sum_{t = 1}^T \nabla_\theta\log\pi_\theta(a_t \lvert s_t)\right)\left(\sum_{t = 1}^T r(\boldsymbol a_t, \boldsymbol s_t)\right)\right],
\end{align*}
where $\tau$ is a trajectory, $\pi_\theta(\tau)$ is the likelihood of executing that trajectory under the current policy $\pi_\theta$, $\pi_\theta(a_t \lvert s_t)$ is the probability of executing action $a_t$ from state $s_t$, and $r(a_t, s_t)$ is the reward gained for that execution.

\subsection{Benchmark for Driving in Dense Traffic}\label{sec:benchmark}

The simulation scenario we consider is implemented using an open-source simulator\footnote{\url{https://github.com/sisl/AutomotiveDrivingModels.jl/}}. In order to obtain diverse on-road behaviours from other vehicles, we make a few modifications to well-known rule-based models - Intelligent Driver Model (IDM) for lane following~\cite{IDM}, and MOBIL for lane changing~\cite{MOBIL}.
IDM is modified to include a \textit{stop-and-go} behaviour that cycles between a non-zero and zero desired velocity in regular time intervals. This behaviour is intended to simulate real-world driving behaviours seen in heavy-traffic during rush-hour. MOBIL is allowed to randomly change lanes (if safe) with probability $p=0.04$. Our benchmark scenario has two key components - dense traffic and a deadend in front of the ego-vehicle. The most important parameters that control an instantiation of this benchmark are the number of vehicles, gaps between them, their desired velocities and the ego-vehicle's distance to the deadend. A detailed list of parameters is given in Table~\ref{tab:simparams}.

\begin{table}[]
\caption{Benchmark scenario parameters}
\label{tab:simparams}
\begin{tabular}{@{}lll@{}}
\toprule
\rowcolor[HTML]{EFEFEF}
\multicolumn{1}{c}{\cellcolor[HTML]{EFEFEF}\textbf{Parameter}} & \multicolumn{1}{c}{\cellcolor[HTML]{EFEFEF}\textbf{Description}} & \multicolumn{1}{c}{\cellcolor[HTML]{EFEFEF}\textbf{Value}} \\ \midrule
$N \in \mathbb{Z}$ & Number of vehicles & $[1, 100]$ \\
$v^{\text{des}} \in \mathbb{R}$ & Desired velocity $(\si{\mPerSecN})$ & $[2, 5]$ \\
$s_0 \in \mathbb{R}$ & Initial gap to vehicle in front $(\si{\meter})$ & $[0.5, 3]$ \\
$s_D \in \mathbb{R}$ & Deadend distance from ego-vehicle $(\si{\meter})$ & $[5, 40]$ \\
$p_c \in \mathbb{R}$ & Cooperativeness & $[0, 1]$ \\
$\lambda_p \in \mathbb{R}$ & Perception range $(\si{\meter})$ & $[-0.15, 0.15]$ \\
$\Delta t$ & Simulation timestep $(\si{\second})$ & $0.2$ \\
$L$ & Number of lanes on road & $\{2, 3\}$ \\
$l$ & Vehicle length $(\si{\meter})$ & 4.0 \\
$w$ & Vehicle width $(\si{\meter})$ & 1.8 \\ \bottomrule
\end{tabular}
\vspace{-10pt}
\end{table}

The cooperativeness and perception range parameters $p_c$ and $\lambda_p$ respectively control whether a vehicle slows down to \textit{cooperate} with another vehicle. Each vehicle on the road can perceive vehicles in its lateral field-of-view, which includes the width of its lane plus an extra width represented by $\lambda_p$.
For any other vehicle that is inside this field-of-view, the vehicle decides whether to slow down, i.e. cooperate, with probability $p_c$ at every timestep $\Delta t$\footnote{We effectively use $p_c = 1$ for vehicles within $w$ of the lateral field-of-view, i.e. enforce full cooperation with vehicles directly in front.}.
In order to elicit complex behaviours from other vehicles on the road that reflect those seen on real roads, we needed to account for different level of cooperativeness ($\lambda_c$), and also the fact that these behaviours vary over time ($p_c$).

\begin{figure}[t]
    \centering
    \begin{subfigure}{\columnwidth}
        \includegraphics[width=\columnwidth]{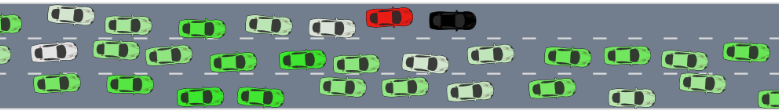}
        \caption{}
        \label{fig:init1}
    \end{subfigure}
    \begin{subfigure}{\columnwidth}
        \includegraphics[width=\columnwidth]{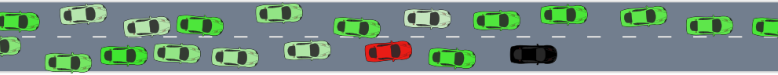}
        \caption{}
        \label{fig:init5}
    \end{subfigure}
    \caption{Randomly generated initial states for the benchmark scenario. The ego-vehicle is in \textcolor{Red1}{red}. Other vehicles are more \textcolor{Green4}{green} if they are more cooperative. We represent the deadend with a black car on the road. (\subref{fig:init1}) Three lane road example. (\subref{fig:init5}) Two lane road example.}
    \label{fig:inits}
    \vspace{-20pt}
\end{figure}

Example initialisations of this benchmark scenario can be seen in Fig.~\ref{fig:inits}.
We colour more cooperative vehicles as more green in the simulation, and less cooperative vehicles as more white. Evaluation code for this scenario is available at \url{https://github.com/dhruvms/DenseTrafficEval/}.

\subsection{Vehicle Model}\label{sec:sim}

All vehicles in our simulation follow a kinematic bicycle model \cite{bicycle}. The nonlinear equations of motion for this model are rewritten here,
\begin{align*}
    \vspace{-5pt}
    \dot{x} &= v \cos(\psi + \beta) \\
    \dot{y} &= v \sin(\psi + \beta) \\
    \dot{\psi} &= \frac{v}{l_r}\sin(\beta) \\
    \dot{v} &= a \\
    \beta &= \arctan\left(\frac{l_r}{l_f + l_r}\tan(\delta_f)\right).
    \vspace{-5pt}
\end{align*}

\begin{figure}[t]
    \centering
    \includegraphics[width=0.5\columnwidth]{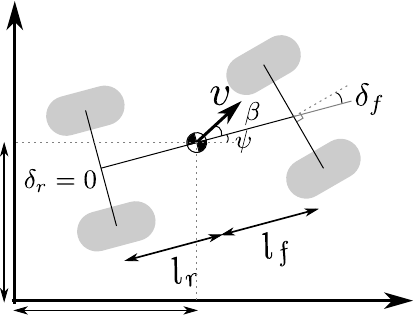}
    \caption{Kinematic bicycle model.}
    \label{fig:bicycyle}
    \vspace{-10pt}
\end{figure}
Relative to a global inertial frame, $(x, y)$ are the spatial coordinates of a vehicle, $\psi$ is the heading, and $v$ is the velocity vector. In the local frame of the vehicle, $\beta$ is the angle of the velocity vector, $\delta_f$ is the angle of the front tires, and $a$ is the acceleration. $l_r$ and $l_f$ are the distances of the rear and front tires respectively from the center of the vehicle. The \textit{steering angle} $\delta_f$ and \textit{acceleration} $a$ are the control inputs for the system. For simplicity, we assume the steering angle of the rear tires $\delta_r = 0$. A diagram of the kinematic bicycle model for four-wheel vehicles can be seen in Fig.~\ref{fig:bicycyle}.

\subsection{Network Architecture and Policy Parameterisation}\label{sec:network}

\begin{figure}[t]
    \centering
    \includegraphics[width=\columnwidth]{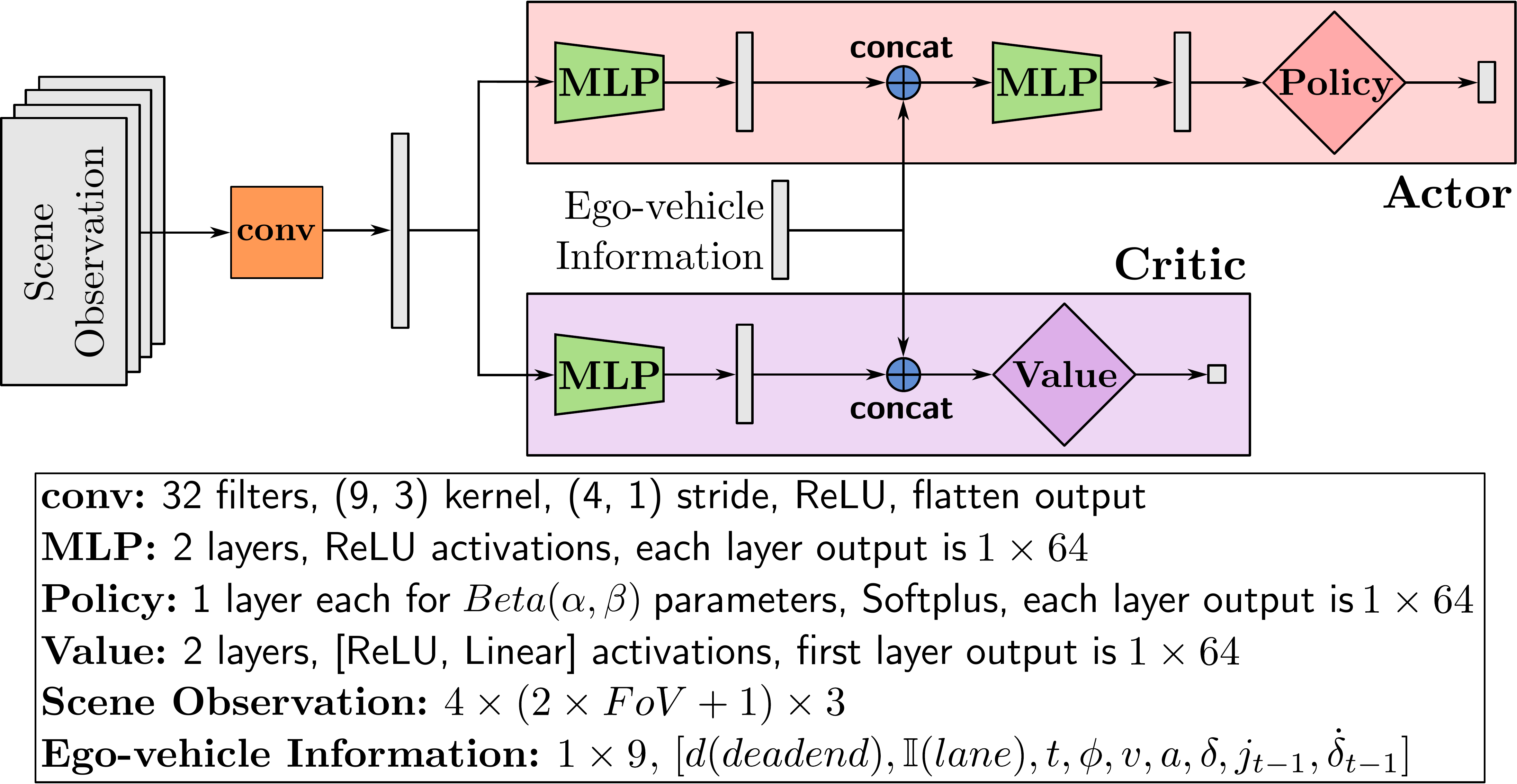}
    \caption{Network architecture.}
    \label{fig:arch}
    \vspace{-10pt}
\end{figure}

We use an actor-critic style network that is trained using Proximal Policy Optimisation (PPO)~\cite{schulman2017proximal}. We found that training was more stable without sharing parameters between the actor and critic networks. Our complete network has around $120,000$ parameters. Fig.~\ref{fig:arch} shows a detailed architecture of our network. The task of autonomous driving is inherently one of continuous control since we usually control the acceleration and steering angle of the vehicle. To get smooth behaviours with high enough fidelity via discrete control would greatly increase the size of the action space, thereby making discrete control methods intractable.
On the contrary, continuous control reinforcement learning has showing promising results in complex problems \cite{lillicrap2015continuous} and even on real-world robots \cite{RLUAV,Hwangboeaau5872}.

For autonomous driving, we also need to keep the comfort of the passengers in mind. Learning a policy over the acceleration and steering angle of a vehicle might lead to jerky or oscillatory behaviour which is undesirable. Instead, we train our network to predict the time derivatives of these quantities, i.e. \textit{jerk} $j$ and steering rate $\dot{\delta}$. This helps us maintain a smooth signal over the true low-level control variables.

Following the analysis in \cite{pmlr-v70-chou17a}, we parameterise our policy as Beta distributions for $j$ and $\dot{\delta}$. This makes training more stable as the policy gradients are unbiased with respect to the finite support of the Beta distribution. We scale each action to acceptable dynamic limits for $j$ and $\dot{\delta}$ inside the simulator. For $j$, we allow values in the range $[-4.0, 2.0]~\si{\mPerSecCuN}$, whereas $\dot{\delta}$ can vary between $[-0.4, 0.4]~\frac{rad}{\si{\second}}$.

\subsection{Ego-vehicle Observations}\label{sec:obs}

Due to the large number of vehicles that could be considered \textit{neighbours} of the ego-vehicle at any time, and the fact that this number would almost certainly change over time, our input representation is agnostic to this number. Additionally, in order to capture the complex inter-vehicle interactions on the road, the input observations include information about the dynamic states of neighbouring vehicles. We use an occupancy-grid style observation that is controlled by one parameter - the longitudinal field-of-view ($FoV$) of the ego-vehicle\footnote{For the experiments in Section~\ref{sec:exps}, $FoV = 50\si{\meter}$ in front and back.}.

\begin{figure}[t]
    \centering
    \includegraphics[width=\columnwidth]{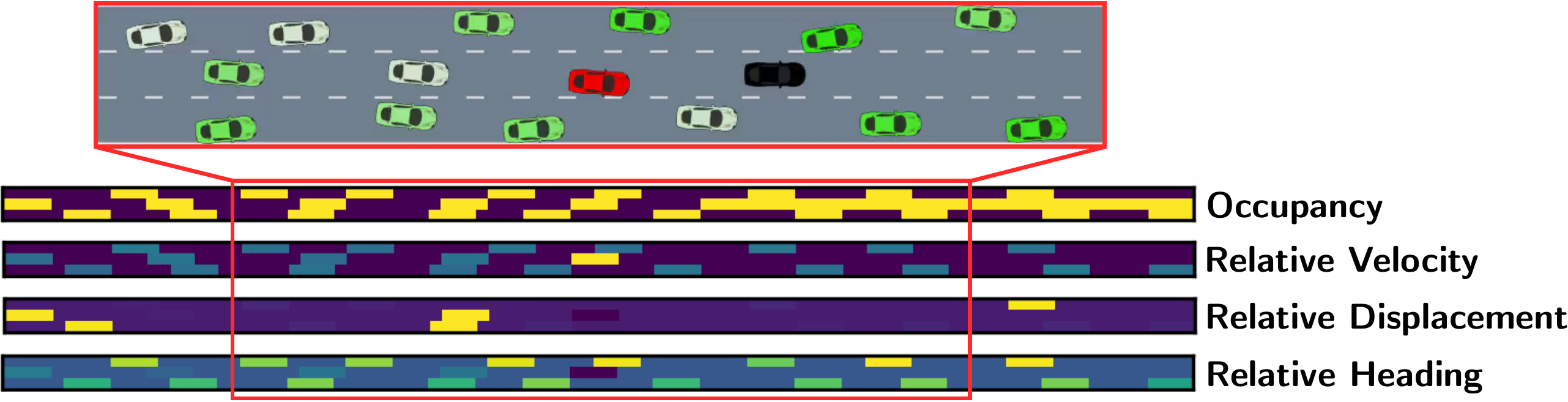}
    \caption{The simulator state (\textit{top}, zoomed in) gets converted to a $4 \times 3 \times (2 \times FoV + 1)$ input observation tensor (\textit{bottom}).}
    \label{fig:obs}
    \vspace{-10pt}
\end{figure}

We assume that in the real-world, on-board sensors and perception systems would process the raw data to determine the relative poses and velocities of neighbouring vehicles. In our simulations, at each timestep, we process the simulator state to calculate an observation tensor of size $4 \times 3 \times (2 \times FoV + 1)$. There is one channel (first dimension) each for on-road occupancy, relative velocities of vehicles, relative lateral displacements, and relative headings with respect to the ego-vehicle. The rows (second dimension) represent the lanes on the road (left, current and right lanes for the ego-vehicle). Fig.~\ref{fig:obs} shows an example of the simulator state and corresponding input observation used for our network.

We also include an ego-vehicle specific feature vector as part of our observation. This includes the distance to the deadend ($d(deadend)$), an indicator for whether we are in the target lane ($\mathbb{I}\{lane\}$), lateral displacement and relative heading from the centerline of the target lane ($t$ and $\phi$), current velocity, acceleration, and steering angle ($v$, $a$, and $\delta$), and the action executed at the last timestep ($j$ and $\dot{\delta}$).

\subsection{Reward Function}\label{sec:reward}
At a high-level, our reward function contains three sets of terms for three different purposes,
\begin{enumerate}[label={\textbf{R\arabic*}}]
    \item We would like the ego-vehicle to be closely oriented with the centerline of the target lane, and travel close to a desired speed.
    \item We want to avoid jerky and oscillatory driving behaviour, since we want to maximise passenger comfort.
    \item Due to an upcoming deadend, we want to finish the lane change maneuver sooner rather than later.
\end{enumerate}
We formulate our reward function by taking these design choices into consideration. Our reward per (state, action) pair is,

\begin{equation*}
\begin{aligned}
    \vspace{-5pt}
    r(\boldsymbol a_t, \boldsymbol s_t) = 0 &- \lambda_v \cdot \lvert v - v_{\text{des}} \lvert \\
    &- \lambda_t \cdot \lvert t \lvert \\
    &- \lambda_\phi \cdot \lvert \phi \lvert \cdot \mathbb{I}\{lane\} \\
    &- \lambda_j \cdot j \\
    &- \lambda_{\dot{\delta}} \cdot \dot{\delta} \\
    &+ 1 \cdot \mathbb{I}\{lane\} \\
    &+ f(deadend)
\end{aligned}
\begin{aligned}
&\left.\vphantom{\begin{aligned}
    - \lambda_v \cdot \lvert v - v_{\text{des}} \lvert \\
    - \lambda_t \cdot \lvert t \lvert \\
    - \lambda_\phi \cdot \lvert \phi \lvert \cdot \mathbb{I}\{lane\}
  \end{aligned}}\right\rbrace\quad\text{\textbf{R1}}\\
&\left.\vphantom{\begin{aligned}
    - \lambda_j \cdot j \\
    - \lambda_{\dot{\delta}} \cdot \dot{\delta}
  \end{aligned}}\right\rbrace\quad\text{\textbf{R2}} \\
&\left.\vphantom{\begin{aligned}
    + 1 \cdot \mathbb{I}\{lane\} \\
    + f(deadend)
  \end{aligned}}\right\rbrace\quad\text{\textbf{R3}}
\end{aligned}
\end{equation*}

$v_{\text{des}}$ is the desired velocity for the ego-vehicle, and $f(deadend)$ rewards or penalises the agent according to ego-vehicle's lane and distance to deadend.


\section{Experimental Results}\label{sec:exps}
\subsection{Baselines}\label{sec:baselines}
We compare the performance of our approach against the following two baselines:
\begin{enumerate}
    \item \emph{MPC baseline~\cite{HowardK07}:} this baseline optimises for the distance of the final state of the trajectory from a target state (in $L_2$-norm distance). All trajectories are $6\si{\second}$ long and the target state is always in the desired lane. We vary three parameters, $s$, $c_f$, and $c_m$ to get the different variants denoted as $\text{MPC}(s, c_f, c_m)$. If the collision check succeeds, we take the first step along the trajectory. Else, we apply full brakes.
    \begin{itemize}
        \item $s \in \{l, 2l, 3l, 4l, 5l\}$ is the longitudinal distance of the target state along the lane\footnote{$l$ is the length of a vehicle in our simulation from Table~\ref{tab:simparams}.}.
        \item $c_f \in \{0, 0.1, 0.25, 0.5, 1.0\}$ represents the fraction of the trajectory we check for collisions with other vehicles ($c = 0$ implies we only check the first state).
        \item $c_m \in \{\text{static}, \text{constant velocity}\}$ is the dynamic model used for other vehicles. $c_m = \text{static}$ treats them as static obstacles, while $c_m = \text{constant velocity}$ propagates them using a constant velocity prediction for the kinematic bicycle model.
    \end{itemize}
    \item \emph{SGAN+MPC~\cite{Sangjae}:} this baseline uses a recurrent neural network to generate predictions~\cite{GuptaJFSA18} for the motions of neighbouring vehicles based on a history of their observations. These predictions are used to create safety constraints for an MPC optimisation that uses Monte Carlo rollouts to compute the (locally) optimal trajectory for the ego-vehicle.
\end{enumerate}
We tested all possible parameter permutations of the former baseline and selected the top two performers for our results in Section~\ref{sec:results}. We also tested the rule-based IDM and MOBIL driving models from Section~\ref{sec:benchmark}. However, these models were not designed with traffic as dense as we consider in mind. As a result, they rarely manage to even initiate a lane-change. For this reason, we do not present numbers for the rule-based baselines in our quantitative evaluation.

\subsection{Evaluation Metrics}\label{sec:metrics}
To evaluate the performance of our approach on the benchmark scenario, we conduct two sets of experiments. In the first experiment (\textbf{E1}), other vehicles on the road do not exhibit the stop-and-go behaviour from Section~\ref{sec:benchmark}. In the second experiment (\textbf{E2}), half the vehicles exhibit stop-and-go behaviours. The simulation is initialised in a way that there will always be vehicles in the desired lane for a preset timeout of $40\si{\second}$. This means that the ego-vehicle must change lanes in traffic, and will not be able to change into the desired lane behind traffic within the timeout limit. We say that an episode ends in \textit{failure} if the ego-vehicle collides with any other vehicle, runs out of room before the deadend, drives off the roadway, or is timed out before successfully changing lanes. We determine \textit{success} if the ego-vehicle successfully changes lanes before the timeout, and stays in the desired lane for at least $5\si{\second}$ without failure.

In addition to the success rate, we quantitatively evaluate performance based on the following metrics (computed only for successful episodes):
\begin{enumerate}[label={\textbf{M\arabic*}}]
    \item \emph{Time to merge:} this is the amount of time elapsed between the start of an episode and successful termination. Lower values are better.
    \item \emph{Minimum distance to other vehicles:} we keep track of the minimum distance of the ego-vehicle to other vehicles on the road for the duration of an episode. Higher values are better.
\end{enumerate}

\subsection{Quantitative Results}\label{sec:results}
\begin{figure}[t]
    \centering
    \includegraphics[width=\columnwidth]{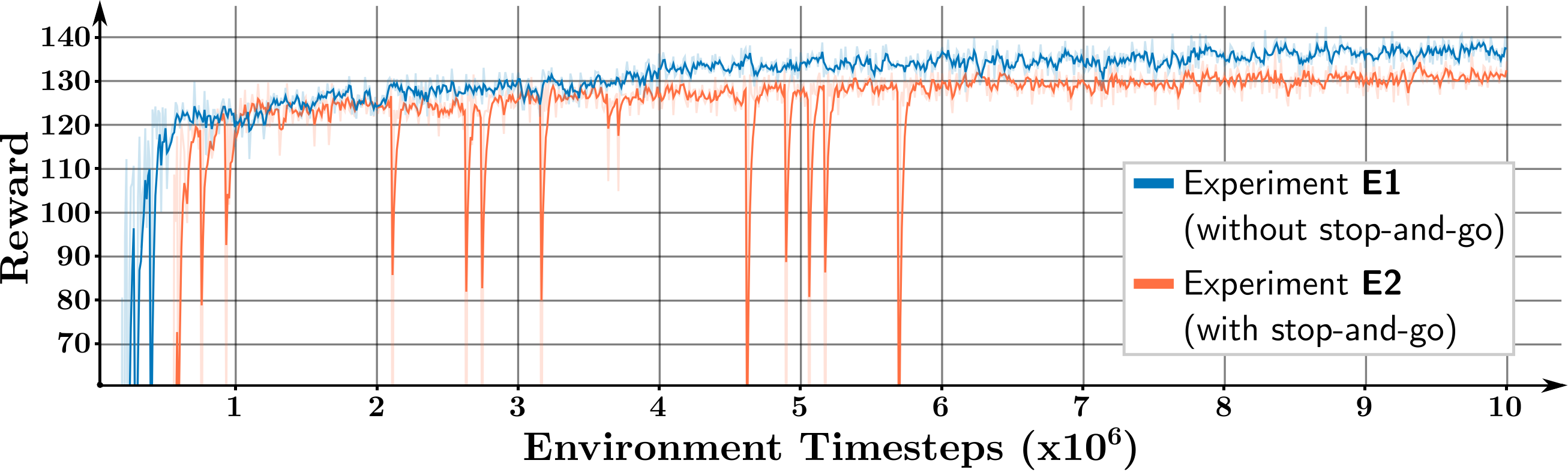}
    \caption{Reward gained during training for experiments \textbf{E1} and \textbf{E2}. Each datapoint is the median reward of the last 10 episodes executed in the environment.}
    \label{fig:rewards}
    \vspace{-10pt}
\end{figure}

We ran all experiments on both two- and three-lane roads, but since they are functionally the same from the ego-vehicle's point-of-view, we only present results for the more complicated three-lane road set of experiments. Tables~\ref{table:exps_nostopgo} and~\ref{table:exps_stopgo} contain results for experiments \textbf{E1} and \textbf{E2} respectively. Fig.~\ref{fig:rewards} shows plots for reward gained during training. For each experiment, for each setting of other drivers on the road, we executed 100 episodes with each model to calculate our statistics. Metrics \textbf{M1} and \textbf{M2} were calculated only from the episodes that finished successfully. We would like to point out that determination of success for SGAN+MPC was on the basis of a more relaxed condition - the egovehicle only had to enter the desired lane, rather than stay in it for $5\si{\second}$ which determined success for the other models.

\begin{table*}[]
\caption{\textbf{Experiment E1}: changing lanes on a road with dense traffic where other drivers do not exhibit random stop-and-go behaviours. Our model far surpasses the performance of MPC baselines. Numbers for this experiment are not available for SGAN+MPC baseline~\cite{Sangjae}.}
\label{table:exps_nostopgo}
\centering
\begin{tabular}{@{}llccc@{}}
\toprule
\multirow{2}{*}{\textbf{Metric}} & \multirow{2}{*}{\textbf{Other Drivers}} & \multicolumn{3}{c}{\textbf{Model}} \\
\cmidrule{3-5}
& & MPC($3.0, 0.5, 2$) & MPC($3.0, 0.25, 2$) & \textbf{Ours} \\ \midrule
Success Rate & Cooperative & 28.43$\%$ & 35.98$\%$ & \cellcolor{PaleGreen1}90$\%$ \\
 & Mixed & 30.61$\%$ & 23.76$\%$ & \cellcolor{PaleGreen1}90.5$\%$ \\
 & Aggressive & 22.73$\%$ & 17.58$\%$ & \cellcolor{PaleGreen1}87$\%$ \\\midrule
Time to merge (\textbf{M1}) & Cooperative & 19.64 ($\pm 11.11$) & 14.33 ($\pm 12.57$) & \cellcolor{PaleGreen1}11.66 ($\pm 5.07$) \\
 & Mixed & 14.78 ($\pm 10.78$) & 11.68 ($\pm 10.74$) & \cellcolor{PaleGreen1}11.10 ($\pm 4.99$) \\
 & Aggressive & 14.72 ($\pm 11.42$) & \cellcolor{PaleGreen1}10.71 ($\pm 9.89$) & 11.51 ($\pm 5.04$) \\\midrule
Minimum distance (\textbf{M2}) & Cooperative & 0.37 ($\pm 0.36$) & 0.36 ($\pm 0.32$) & \cellcolor{PaleGreen1}0.38 ($\pm 0.23$) \\
 & Mixed & 0.39 ($\pm 0.29$) & \cellcolor{PaleGreen1}0.43 ($\pm 0.40$) & 0.32 ($\pm 0.21$) \\
 & Aggressive & \cellcolor{PaleGreen1}0.42 ($\pm 0.41$) & 0.31 ($\pm 0.31$) & 0.30 ($\pm 0.19$)) \\ \bottomrule 
\end{tabular}
\end{table*}

\begin{table*}[]
\caption{\textbf{Experiment E2}: other drivers on the road exhibit random stop-and-go behaviours. MPC baselines do well in terms of \textbf{M1} and \textbf{M2}, but they have abysmal success rates. Our model performs well across the board, outperforming SGAN+MPC on the important metric \textbf{M1}. Note that SGAN+MPC uses a relaxed criterion for success.}
\label{table:exps_stopgo}
\centering
\begin{tabular}{@{}llcccc@{}}
\toprule
\multirow{2}{*}{\textbf{Metric}} & \multirow{2}{*}{\textbf{Other Drivers}} & \multicolumn{4}{c}{\textbf{Model}} \\
\cmidrule{3-6}
& & MPC($3.0, 0.5, 2$) & MPC($3.0, 0.25, 2$) & SGAN+MPC~\cite{Sangjae} & \textbf{Ours} \\ \midrule
Success Rate & Cooperative & 23.44$\%$ & 25.95$\%$ & \cellcolor{PaleGreen1}99$\%$ & 85.5$\%$ \\
 & Mixed & 21.43$\%$ & 21.62$\%$ & \cellcolor{PaleGreen1}97$\%$ & 87$\%$ \\
 & Aggressive & 20.21$\%$ & 20.63$\%$ & \cellcolor{PaleGreen1}81$\%$ & 80$\%$ \\\midrule
Time to merge (\textbf{M1}) & Cooperative & 16.54 ($\pm 12.44$) & \cellcolor{PaleGreen1}14.69 ($\pm 12.82$) & 23.97 ($\pm 5.10$) & 16.40 ($\pm 9.07$) \\
 & Mixed & 19.88 ($\pm 11.31$) & \cellcolor{PaleGreen1}8.43 ($\pm 8.46$) & 25.76 ($\pm 5.37$) & 18.02 ($\pm 9.58$) \\
 & Aggressive & 14.54 ($\pm 10.98$) & \cellcolor{PaleGreen1}8.71 ($\pm 8.26$) & 29.32 ($\pm 7.07$) & 17.99 ($\pm 9.94$) \\\midrule
Minimum distance (\textbf{M2}) & Cooperative & \cellcolor{PaleGreen1}0.50 ($\pm 0.39$) & 0.38 ($\pm 0.40$) & 0.49 ($\pm 0.18$) & 0.38 ($\pm 0.24$) \\
 & Mixed & \cellcolor{PaleGreen1}0.44 ($\pm 0.39$) & 0.32 ($\pm 0.24$) & 0.42 ($\pm 0.19$) & 0.35 ($\pm 0.23$) \\
 & Aggressive & 0.38 ($\pm 0.27$) & \cellcolor{PaleGreen1}0.39 ($\pm 0.38$) & 0.30 ($\pm 0.14$) & 0.37 ($\pm 0.23$)) \\ \bottomrule 
\end{tabular}
\vspace{-10pt}
\end{table*}


For the experiments in Tables~\ref{table:exps_nostopgo} and~\ref{table:exps_stopgo}, there were $60$ vehicles on the road, and the front-to-tail gap between two vehicles was randomly sampled to be between $[0.5, 3]\si{\meter}$ (for reference each car is $4\si{\meter}$ long). There are several observations to be made on the basis of the numbers from Tables~\ref{table:exps_nostopgo} and~\ref{table:exps_stopgo}.
\begin{enumerate}[label={\textbf{O\arabic*}}]
    \item The MPC baselines have poor success rates since they fail to account for any inter-vehicle interactions.
    \begin{itemize}
        \item The \textit{static} collision checking scheme performed much worse than \textit{constant velocity}, as expected.
        \item These baselines suffer from both too small $c_f$ values (ego-vehicle is aggressive which causes collisions) and too large $c_f$ values (ego-vehicle is pessimistic which causes freezing).
    \end{itemize}
    \item The MPC models, if successful, do well in terms of \textbf{M1} and \textbf{M2} as they minimise distance to a target point which makes them speed up quickly, but also brake hard if a computed trajectory might collide.
    \item SGAN+MPC achieves remarkable success rates (albeit with a relaxed criterion) on the benchmark task. It is also affected by the types of other drivers on the road as performance degrades with increasing numbers of aggressive drivers.
    \item Our model arguably performs the best out of all baselines across the board.
    \begin{itemize}
        \item In the worst-case, we achieve $80\%$ success rate on the difficult benchmark scenario with stop-and-go behaviours, with a stricter criterion for success.
        \item Our times to merge (\textbf{M1}) are much lower than SGAN+MPC partly because our model does not need to wait for observations of neighbouring vehicles to generate predictions for their motion.
        \item We have comparable performance to SGAN+MPC on \textbf{M2}.
        \item Our model-free approach is not affected by the distribution of drivers on the road.
    \end{itemize}
\end{enumerate}

\begin{figure}[t]
    \centering
    \begin{subfigure}{\columnwidth}
        \includegraphics[width=\columnwidth]{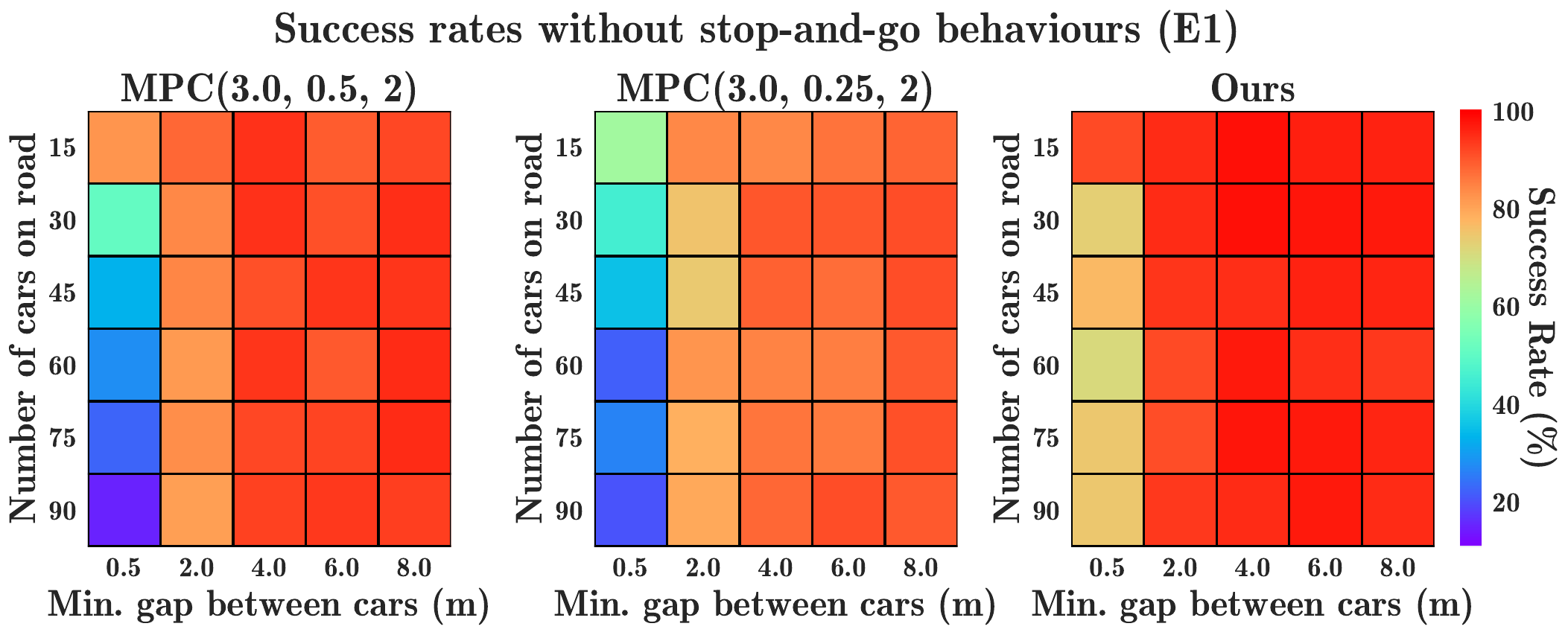}
        \caption{}
        \label{fig:no_sg}
    \end{subfigure}
    \begin{subfigure}{\columnwidth}
        \includegraphics[width=\columnwidth]{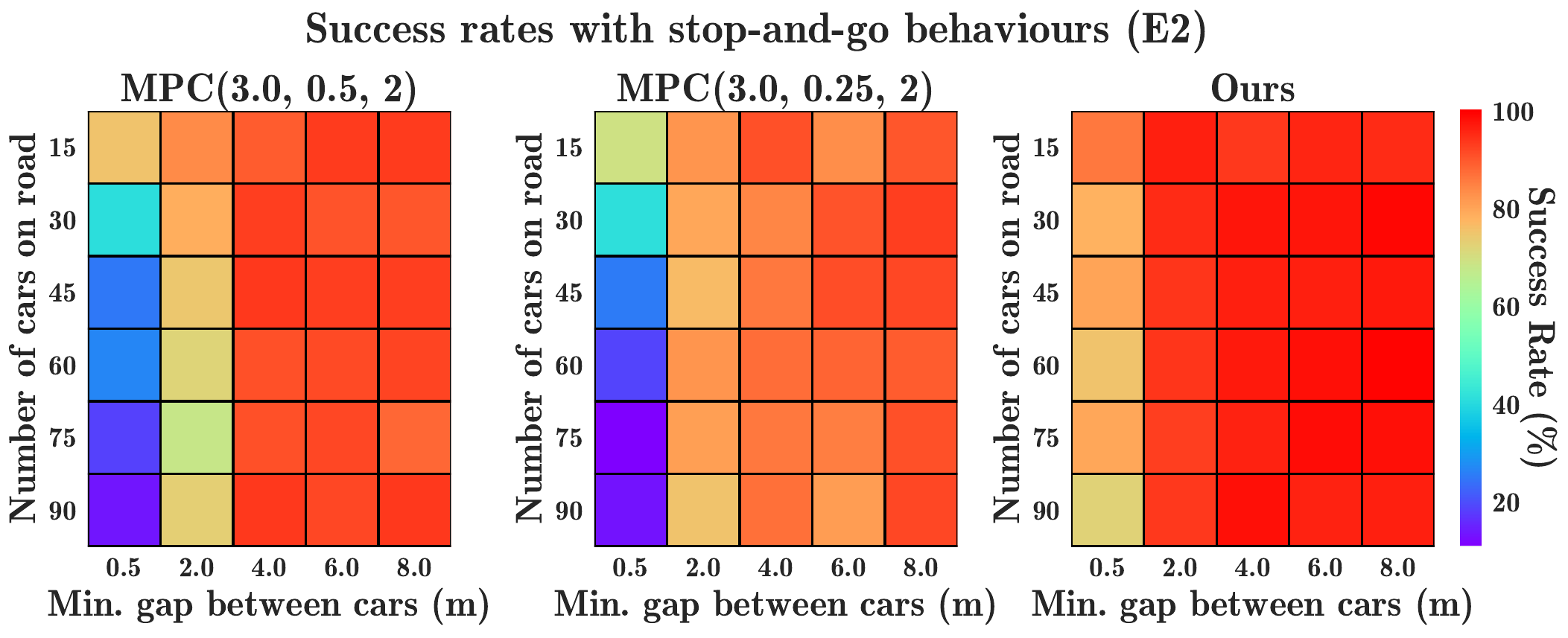}
        \caption{}
        \label{fig:sg}
    \end{subfigure}
    \caption{Comparison of model performance with varying numbers of vehicles on the road and gaps between them.}
    \label{fig:cars_gaps}
    \vspace{-10pt}
\end{figure}

In order to demonstrate the robustness and generalisation capability of our approach, we varied the number of vehicles on the road, and also the gaps between them. For this experiment, all other drivers on the road were assigned random cooperativeness parameters $p_c$ (this corresponds to the \textit{Mixed} scenario for \textbf{E1} and \textbf{E2}). Since numbers for such an experiment are not available for SGAN+MPC, we only compare against our MPC baselines. Fig.~\ref{fig:cars_gaps} shows the results of this experiment pictorially. Each image is a heatmap of success rates for different numbers of vehicles and gaps, both without (Fig.~\ref{fig:no_sg}) and with (Fig.~\ref{fig:sg}) stop-and-go behaviours. Our model performs much better than either baseline in terms of success rates. Moreover, our model's performance only degrades slightly with increasing complexity of the scenario. Across the $60$ different experiments in Fig.~\ref{fig:cars_gaps}, only in \textit{one} case ($60$ vehicles, $8.0\si{\meter}$ gap, no stop-and-go) did a baseline, MPC$(3.0, 0.5, 2)$, do better than our model ($95\%$ versus $93.5\%$ success rate). Our code can be found at \url{https://github.com/dhruvms/HighwayTraffic}.

\section{Conclusion}\label{sec:conc}
We present a benchmark scenario for the task of driving and merging in dense traffic.
This requires complex decision-making algorithms that need to reason about interactions with neighbouring vehicles, implicitly or explicitly, in order to successfully accomplish the task.
Our model-free approach does not rely on driver models of other vehicles, or even on predictions about their motions, and is successful on the benchmark. It outperforms traditional rule-based models (which fail entirely) and model-predictive control based models. It also performs comparably to a new approach~\cite{Sangjae} in terms of success rates, and does better on key metrics, without using predictions about trajectories of other vehicles.

Since we blindly execute our policy, we cannot guarantee collision free execution. Even in the worst-case, we cannot guarantee that the autonomous vehicle will remain stationary if all actions are dangerous. For this reason, we would like to investigate the use of an \textit{overseer} that determines whether an action predicted by our policy is safe to execute. This could also be done by training the policy to predict target states for an MPC-like controller, thereby offloading execution to traditional methods than can easily incorporate complex constraints.
Making our approach more model-based by incorporating either some minimal information about other driver models, or a notion of the desired effect of interacting with them within the reinforcement learning framework is another promising direction of research.




\section*{Acknowledgement}
The authors thank David Isele at Honda Research Institute USA, Inc., and Huckleberry Febbo and Ran Tian at University of Michigan, Ann Arbor, USA for helpful discussions during this work. Dhruv Mauria Saxena completed this work during his internship at Honda Research Institute USA, Inc.



\balance
{\small
\bibliographystyle{plain}
\bibliography{references}
}

\end{document}